# Advancing Harmful Content Detection in Organizational Research: Integrating Large Language Models with Elo Rating System

**Mustafa Akben**　　**Aaron Satko**
Elon University, NC　Elon University, NC

## Abstract

*Large language models (LLMs) offer promising opportunities for organizational research. However, their built-in moderation systems can create problems when researchers try to analyze harmful content, often refusing to follow certain instructions or producing overly cautious responses that undermine validity of the results. This is particularly problematic when analyzing organizational conflicts such as microaggressions or hate speech. This paper introduces an Elo rating-based method that significantly improves LLM performance for harmful content analysis In two datasets, one focused on microaggression detection and the other on hate speech, we find that our method outperforms traditional LLM prompting techniques and conventional machine learning models on key measures such as accuracy, precision, and F1 scores. Advantages include better reliability when analyzing harmful content, fewer false positives, and greater scalability for large-scale datasets. This approach supports organizational applications, including detecting workplace harassment, assessing toxic communication, and fostering safer and more inclusive work environments.*

## 1. Introduction

Organizations often struggle to navigate and address harmful communication and content at work, characterized by abusive language, microaggressions, and hate speech and thus creates a hostile, discriminatory, or unsafe work environment (Cortina et al., 2018; Smith & Griffiths, 2022; Tuckey & Neall, 2014a). Harmful content can be shared on internal message boards, in emails from supervisors, or on organization social media platforms. Such harmful communication might decrease employee performance (Lim & Cortina, 2005), increase turnover (Jones et al., 2016), and harm employee well-being (Smith & Griffiths, 2022). Furthermore, it can damage the organization's reputation, thereby negatively influencing overall performance (Lutgen-Sandvik & Tracy, 2012). Therefore, developing methods to identify and study harmful content is key for preventing adverse work outcomes and advancing organizational science.

Traditionally, researchers studying harmful content have relied on self- or other-rated survey measures (Smith & Griffiths, 2022). While useful in many contexts, these methods are often limited by social desirability biases or by the raters' interpretation of the content (Podsakoff et al., 2003), which can compromise the reliability and accuracy of the study (Krippendorff, 1970). However, traditional machine learning methods are limited by their reliance on large training datasets and a lack of semantic understanding (i.e., understanding the meaning of written text beyond character representations) (Chollet, 2021). Recently, however, large language models (LLMs) have opened a promising new avenue for social science research (Ziems et al., 2024). LLMs are effective at processing and understanding textual data, attending to both syntactic aspects of the text (i.e., word choices and grammatical structure) and semantic aspects (i.e., meaning and implication) (Brown et al., 2020; Devlin et al., 2019). Their pre-training allows them to effectively process text with little to no data, whereas doing so previously required massive datasets to train conventional machine learning algorithms (Kaplan et al., 2020). This enables LLMs to perform advanced NLP tasks, in which they often outperform their predecessors (Qin et al., 2024).

However, LLMs also face notable challenges. When tasked with detecting harmful content, for instance, they often underperform (Cui et al., 2024). Rather than reflecting a fundamental lack of capability, this issue largely stems from their internal content moderation mechanisms (Xie et al., 2024). While these moderation systems are necessary for ethical and responsible use, they can impair the analysis of harmful content by either refusing to process the requests or by generating overly sanitized outputs, often resulting in subpar accuracy (Solaiman, 2023). Utilizing cutting-edge LLM systems in organizational and social sciences calls for new methods for examining harmful content while respecting the model's existing content moderation.

To address the challenges, we propose a new methodological approach that integrates large language models (LLMs) with the Elo rating algorithm (Elo, 1978), which was initially designed to rank chess players and later applied in fields like sports science (Hvattum & Arntzen, 2010) and machine learning (Silver et al., 2018). In our procedure, the LLM operates in a "tournament" format, making pairwise comparisons between texts to determine which exhibits the harmful content more prominently. Instead of generating harmful content or lengthy justifications for it, the LLM provides a binary choice, thereby minimizing risks related to content moderation. We then apply the Elo algorithm to these pairwise judgments with a logistic transformation to convert them into probability-based scores that reflect the likelihood that each text contains harmful content relative to others. To validate our approach, we tested it on two distinct datasets: one consisting of microaggressions and the other of hate speech, thus exhibiting two different spectra of harmfulness. For comparison, we benchmarked our method against traditional machine learning models (i.e., random forest, support vector machine, gradient boosting machines, and logistic regression) and common LLM prompting techniques (e.g., zero-shot prompting). Our method obtained higher classification accuracy, precision, recall, F1 scores, and balanced accuracy, suggesting better reliability and statistical power improvement.

## 2. Literature Review

### 2.1 Traditional Machine Learning Methods

Machine learning (ML) has proven to be a faster, less labor-intensive method of text classification than manual coding (Duriau et al., 2007; Grimmer & Stewart, 2013). These methods learn associations between inputs (texts) and outputs (classes or categories) via supervised text classification (Chollet, 2021). Supervised text classification involves training algorithms on labeled datasets, pairing each text with a category such as "harmful" or "not harmful." Algorithms like random forests, support vector machines, and gradient boosting machines learn to identify patterns between texts' labeled categories and their extracted features (Joachims, 1998). Features may involve extracting word frequencies, n-grams, text vectorization, or term frequency-inverse document frequency (TF-IDF). After being trained, these models classify new data by leveraging patterns identified in training datasets. When trained on large and representative datasets, supervised learning achieves high accuracy (Kowsari et al., 2019).

Despite their success, traditional supervised ML approaches are often ill-suited for classifying harmful content in social science research. Such approaches require large, labeled datasets, often unavailable or too costly to create for nuanced categories such as microaggressions or coded hate speech (Schmidt & Wiegand, 2017; Wiegand et al., 2018). Manual labeling is time-intensive and prone to biases and subjectivity. In addition, these models struggle with semantic complexities, such as sarcasm, irony, and contextual subtleties, which are critical for detecting harmful content (Davidson et al., 2017). Social science studies tend to have relatively small sample sizes, significantly degrading the performance of data-hungry ML models, causing overfitting and poor generalization to new data (Ying, 2019). This challenge is particularly relevant for detecting harmful content because models may fail to detect rare or emerging harmful expressions. Thus, supervised ML's limitations regarding data requirements, interpretability, and semantic complexity necessitate careful consideration—and perhaps alternative approaches—when studying harmful content in organizational settings.

## 2.2 Large Language Models

Large language models (LLMs) represent major advances in natural language processing (NLP), offering new capabilities for understanding and producing human-like text. LLMs, such as OpenAI's GPT models (Brown et al., 2020; Radford et al., 2019), Google's BERT model (Devlin et al., 2019), and Meta's LLaMA model (Touvron et al., 2023), use a deep learning architecture referred to as the transformer network. Due to the scale of these models with billions or trillions of parameters, they can complete a wide array of NLP tasks, often with minimal task-specific fine-tuning (Brown et al., 2020).

In the social sciences, LLMs have become versatile research tools. LLMs can perform at state-of-the-art levels on NLP benchmarks (Ziems et al., 2024), especially in low-resource settings with minimal training data. They often surpass traditional machine-learning models (Luo et al., 2024). Moreover, their capacity to learn from small (few-shot prompting) or no data at all (zero-shot prompting) enables LLMs to be adapted to diverse research questions in the social sciences (Brown et al., 2020). Researchers have used LLMs to investigate political polarization in social media (Li et al., 2024; Zeng et al., 2024) and predict mental health issues based on online posts (Guntuku et al., 2019). Moreover, LLMs have surpassed the performance of human judges on crowdsourcing platforms (such as MTurk) in accuracy for text classification tasks (Gilardi et al., 2023).

However, the implementation of such models in detecting harmful content exposes a critical dilemma between efficacy and ethics. While LLMs demonstrate impressive abilities they often underperform when assessing harmful text. This issue is driven partly by content moderation systems already hardwired into LLMs designed to filter harmful material (Gillespie, 2018). This tradeoff often yields high rates of false positives and false negatives (Weidinger et al., 2022). Moderation systems often err on the side of caution, causing the model to refuse seemingly innocuous prompts. For example, Cui et al. (2024) found that in the OR-Bench task, assessing LLMs' tendencies to over-refuse seemingly, but not actually toxic prompts, over-rejection rates were as high as 74%.

Another significant challenge arises from prompt engineering, which involves writing clear instructions to elicit desired outcomes from an LLM (Wei et al., 2022). If prompts are poorly structured, the model may produce inconsistent responses (Reynolds & McDonell, 2021), resulting in a higher probability of false positives. Two widespread techniques include zero-shot prompting, instructing the model without examples (Brown et al., 2020), and chain-of-thought prompting, directing the model through a sequential reasoning process—often generated by the model itself (Wei et al., 2022). Although chain-of-thought prompting can improve performance, it can also heighten the risk of triggering filtering mechanisms by revealing intermediate steps. Thus, the model may withhold a response altogether or simply provide a canned, unhelpful response, limiting its usefulness where analysis of potentially harmful material may be imperative.

## 3. Materials and Methods

### 3.1 The Elo Rating Explained

The Elo rating system is an adaptive, relative ranking system measuring player skill in zero-sum games (Elo, 1978). Elo scores for each agent result from comparisons with other agents in competitive interactions. The Elo method mirrors actual human comparative judgments, underlying its broad application beyond chess (Hvattum & Arntzen, 2010). Elo converts results of interactions into scores reflecting relative skill, continuously evaluating expected success. The Elo score implements a probability model with a logistic transformation. Specifically, if the Elo score for player A is RA and the Elo score for player B is RB, the probability that player A will defeat player B is:

$$(1) \quad P(A) = \frac{1}{\left(1 + 10^{\frac{R_B - R_A}{\beta}}\right)}$$

The constant β is a scaling parameter whereby larger values of beta correspond to smaller adjustments in player scores after each match. β is often set to 400. The formulation in equation (1) represents a logistic curve bounded between 0 and 1, whereby the probability of winning increases as the difference between the scores increases. The probability, P(B), that player B wins over player A is P(B) = 1 − P(A).

After each game, ratings are updated based on the observed versus expected outcome:

$$(2) \quad R'_A = R_A + K(S_A - P(A))$$

where $R'_A$ represents the new rating of player $A$, $K$, modulates rating sensitivity changes and $S_A$ corresponds to the results of the comparison (1 for a win, 0.5 for a draw, and 0 for a loss). Ratings adjust significantly for unexpected outcomes and minimally for expected outcomes.

## 3.2 Proposed Method

When paired with large language models (LLMs), Elo ratings offer key advantages for harmful content detection. The comparative nature of Elo shifts the focus from absolute judgments to pairwise comparisons, aligning with human decision-making. Elo aggregates comparisons into continuous probability-based ratings, reducing binary classification limitations. Its adaptability allows incremental incorporation of new data without retraining. Using LLMs only for pairwise comparisons minimizes content moderation risks, enabling ethical investigation without violating moderation policies (Cui et al., 2024).

Our proposed approach brings together the strength of large language models (LLMs) and

the statistical foundation of the Elo rating system in three steps:

### 3.2.1 Step 1. Tournament Construction.
Following tournament design principles (Bradley & Terry, 1952), classification is treated as pairwise comparisons between N entries. For computational efficiency, we use a balanced incomplete block design (Montgomery, 2017), comparing each text to m randomly selected texts from the dataset where m < N. Due to Elo's transitivity principle, incomplete designs preserve ranking accuracy with fewer comparisons.

### 3.2.2 Step 2. LLM-Based Comparison.
Large language models (LLMs) make paired comparisons using simplified comparative questions. For each entry pair (A, B), the LLM identifies the text exhibiting stronger harmful content, returning only "A" or "B" to avoid moderation triggers.

### 3.2.3 Step 3. Elo Rating Calculation and Probability Transformation.
We recalibrate Elo ratings based on LLM-derived comparisons and convert ratings to probabilities using:

$$(3) \quad P(A_n) = \frac{1}{e^{\frac{R_{A_n}}{c}}}$$

where c is a scaling factor ensuring probability distribution, and normalized Elo rating ranges from 0 to 1, reflecting relative harmfulness.

## 3.3 Datasets

We evaluated our model on two datasets representing different harmful communication severities. Accuracy metrics compared our method against supervised machine learning systems (support vector machines, random forests, gradient boosting machines, logistic regression) and zero-shot large language model prompting.

### 3.3.1 Microaggression Dataset.
To assess the proposed method's effectiveness in addressing nuanced harmful content with lower linguistic markers, we created a new dataset. Our original dataset came from a public machine learning competition and contained a series of workplace emails to employees (N=1000), but it had no microaggression content. To introduce microaggressions, we inserted carefully crafted sentences into randomly selected data points (N=500) that contained various forms of microaggressions described in the literature (Sue & Spanierman, 2020). These randomly selected data points became our microaggression class. These included verbal, behavioral, and structural microaggressions, which are known to capture all forms of microaggressions. After this generation process, two independent raters who were blind to the process were asked whether each data point contained microaggression. The interrater agreement, calculated using Cohen's κ, was greater than the recommended cutoff of .75, indicating strong agreement (McHugh, 2012).

**3.3.2 Hate Speech Dataset.** For analyzing harmful content that has a greater intensity and clearer linguistic markers, we utilized a publicly available hate speech dataset comprising 24,802 entries (Davidson et al., 2017). This dataset encompasses a wide range of hate speech categories, including but not limited to racial, religious, ethnic, gender, and sexual orientation-based hate speech. The inclusion of such diverse categories enables a comprehensive examination of hate speech manifestations in various contexts and forms.

# 4. Findings

## 4.1 Microaggression Detection

Table 1 reports the omnibus tests for the microaggression detection task, which reveal significant differences among the compared methods on all examined metrics. Specifically, both the Kruskal-Wallis and ANOVA tests were statistically significant across accuracy, F1 score, Matthews Correlation Coefficient (MCC), precision, recall, balanced accuracy, and ROC AUC ($p < .001$). Table 2 provides a detailed breakdown of the mean performance (and standard deviations), while Figure 1 illustrates the comparative accuracy across all methods and Figure 2 depicts the ROC curves.

Our proposed method achieved the highest overall classification results, with an accuracy of .81 (SD = .03), an F1 score of .82 (SD = .03), and an MCC of .62 (SD = .05). In addition, the model demonstrated both high precision (.82, SD = .06) and recall (.82, SD = .09). These metrics reflect a balanced capacity to correctly identify microaggressions without overly misclassifying benign statements. Notably, the proposed approach also attained a robust ROC AUC of .88 (SD = .02), indicating strong discriminative performance.

### Table 1. ANOVA and Non-Parametric Test Results for Microaggression Detection Task

| Performance Metric | Kruskal-Wallis Test | ANOVA Test |
|---|---|---|
| Accuracy | $\chi^2$ (6) = 264.56, p < .001 | F (6, 693) = 69.63, p < .001 |
| F1 Score | $\chi^2$ (6) = 316.83, p < .001 | F (6, 693) = 103.90, p < .001 |
| MCC | $\chi^2$ (6) = 295.72, p < .001 | F (6, 693) = 73.89, p < .001 |
| Precision | $\chi^2$ (6) = 398.49, p < .001 | F (6, 693) = 283.30, p < .001 |
| Recall | $\chi^2$ (6) = 269.05, p < .001 | F (6, 693) = 137.70, p < .001 |
| Balanced Accuracy | $\chi^2$ (6) = 270.79, p < .001 | F (6, 693) = 69.88, p < .001 |
| ROC AUC | $\chi^2$ (6) = 334.75, p < .001 | F (6, 693) = 108.00, p < .001 |

*Note.* Data distributions for some metrics were skewed, necessitating the use of both parametric (ANOVA) and non-parametric (Kruskal-Wallis) tests. $\chi^2$ values represent the results of the Kruskal-Wallis's test, and *F* values represent the results of ANOVA. Degrees of freedom (df) are reported in parentheses. Performance metrics evaluated include accuracy, balanced accuracy, F1 score, precision, recall, Matthews Correlation Coefficient (MCC), and ROC AUC (Area Under the Curve). Bootstrapping was employed with two hundred samples per iteration, repeated 100 times per dataset, to reflect data scarcity conditions often encountered in social science research.

### Table 2. Classification Performance in Predicting Microaggressions

| Model | Accuracy | F1 Score | MMC | Precision | Recall | Balanced Accuracy | ROC |
|---|---|---|---|---|---|---|---|
| Proposed Model | **.81 (.03)** | **.82 (.03)** | **.62 (.05)** | .82 (.06) | .82 (.09) | **.80 (.03)** | **.88 (.02)** |
| LLM with Zero-shot Prompt | .73 (.03) | .66 (.04) | .55 (.04) | **.98 (.02)** | .50 (.05) | .75 (.03) | .75 (.03) |
| Random Forest | .73 (.04) | .76 (.05) | .48 (.08) | .71 (.05) | .84 (.11) | .72 (.04) | .77 (.05) |
| Support Vector Machine | .72 (.05) | .74 (.05) | .44 (.09) | .73 (.07) | .77 (.11) | .71 (.05) | .75 (.06) |
| Gradient Boosting | .72 (.05) | .76 (.06) | .45 (.10) | .71 (.07) | **.84 (.12)** | .71 (.06) | .75 (.07) |
| Logistic Regression | .73 (.04) | .75 (.05) | .46 (.08) | .73 (.06) | .79 (.11) | .72 (.04) | .77 (.05) |

*Note.* Results are presented as mean (standard deviation) based on one hundred simulations. In each simulation, two hundred data points were randomly sampled from a dataset. MCC = Matthews Correlation Coefficient; ROC AUC = Receiver Operating Characteristic Area Under the Curve; LLM = Large Language Model.

Compared to traditional supervised learning methods and zero-shot LLM prompting, the proposed method maintained its superiority across all indices. In contrast, zero-shot prompting recorded exceptionally high precision (.98, SD = .02) but relatively low recall (.50, SD = .05), resulting in only moderate F1 performance. Meanwhile, the machine learning benchmarks (e.g., random forest, support vector machines, gradient boosting, and logistic regression) obtained similar outcomes, with accuracy values between .72 and .73. However, they generally evidenced lower MCC values than the proposed framework, underscoring the advantage of pairing an LLM with an Elo-based approach. Post-hoc comparisons, corrected for multiple testing, showed that the proposed method outperformed all other models at p < .05, at accuracy, F1, balanced accuracy and ROC scores.

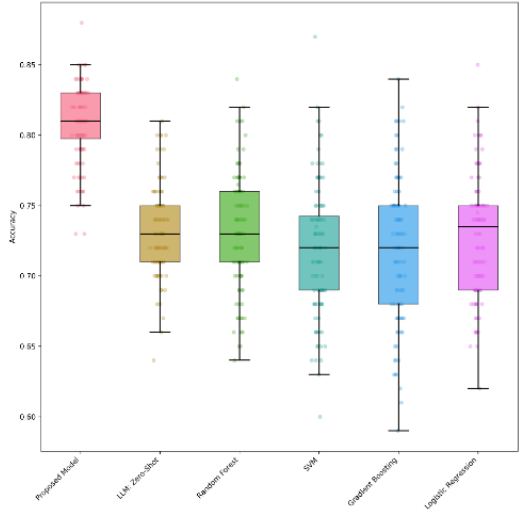

**Figure 1. Accuracy Comparisons for the Microaggression Classification Task**

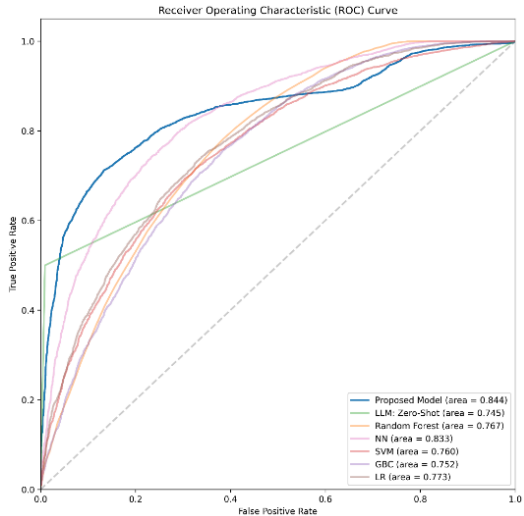

**Figure 2. ROC Curve for the Microaggression Classification Task**

### 4.2 Hate Speech Detection

A parallel pattern was observed in the hate speech detection task (Table 3), where omnibus tests again identified statistically significant performance differentials across all metrics (p < .001). As detailed in Table 4, the proposed method once more led in overall effectiveness. It achieved an accuracy of .86 (SD = .03), along with an F1 score of .90 (SD = .02) and an MCC of .67 (SD = .06). Of particular note, it recorded a recall of .95 (SD = .04), suggesting it adeptly captured a broad spectrum of hate speech variations. Figures 3 and 4 respectively present the accuracy measures relative to other methods and the ROC curves for this dataset.

### Table 3. ANOVA and Non-Parametric Test Results for Hate Speech Detection Task

| Performance Metric | Kruskal-Wallis Test | ANOVA Test |
|---|---|---|
| Accuracy | $\chi^2$ (6) = 240.49, p < .001 | F (6, 693) = 64.04, p < .001 |
| F1 Score | $\chi^2$ (6) = 238.74, p < .001 | F (6, 693) = 59.44, p < .001 |
| MCC | $\chi^2$ (6) = 248.09, p < .001 | F (6, 693) = 62.87, p < .001 |
| Precision | $\chi^2$ (6) = 229.76, p < .001 | F (6, 693) = 59.11, p < .001 |
| Recall | $\chi^2$ (6) = 173.53, p < .001 | F (6, 693) = 59.85, p < .001 |
| Balanced Accuracy | $\chi^2$ (6) = 223.00, p < .001 | F (6, 693) = 53.77, p < .001 |
| ROC AUC | $\chi^2$ (6) = 240.65, p < .001 | F (6, 693) = 61.43, p < .001 |

*Note.* Data distributions for some metrics were skewed, necessitating the use of both parametric (ANOVA) and non-parametric (Kruskal-Wallis) tests. $\chi^2$ values represent the results of the Kruskal-Wallis's test, and *F* values represent the results of ANOVA. Degrees of freedom (df) are reported in parentheses. Performance metrics evaluated include accuracy, balanced accuracy, F1 score, precision, recall, Matthews Correlation Coefficient (MCC), and ROC AUC (Area Under the Curve). Bootstrapping was employed with two hundred samples per iteration, repeated 100 times per dataset, to reflect data scarcity conditions often encountered in social science research.

### Table 4. Classification Performance in Predicting Hate Speech

| Model | Accuracy | F1 Score | MMC | Precision | Recall | Balanced Accuracy | ROC |
|---|---|---|---|---|---|---|---|
| Proposed Model | **.86 (.03)** | **.90 (.02)** | **.67 (.06)** | .85 (.05) | **.95 (.04)** | **.81 (.04)** | **.88 (.03)** |
| LLM with Zero-shot Prompt | .78 (.04) | .82 (.04) | .55 (.07) | **.88 (.04)** | .79 (.09) | .78 (.04) | .80 (.05) |
| Random Forest | .76 (.04) | .83 (.03) | .42 (.12) | .78 (.05) | .89 (.06) | .69 (.06) | .76 (.06) |
| Support Vector Machine | .80 (.04) | .85 (.03) | .51 (.10) | .83 (.04) | .89 (.06) | .75 (.06) | .83 (.05) |
| Gradient Boosting | .76 (.05) | .84 (.03) | .44 (.14) | .78 (.05) | .91 (.07) | .69 (.08) | .77 (.06) |
| Logistic Regression | .79 (.04) | .85 (.03) | .52 (.10) | .82 (.05) | .90 (.06) | .74 (.06) | .83 (.05) |

*Note.* Results are presented as mean (standard deviation) based on one hundred simulations. In each simulation, two hundred data points were randomly sampled from a dataset. MCC = Matthews Correlation Coefficient; ROC AUC = Receiver Operating Characteristic Area Under the Curve; LLM = Large Language Model.

Zero-shot LLM prompting yielded an accuracy of .78 (SD = .04) and an F1 score of .82 (SD = .04), indicating notable performance but trailing behind the proposed approach on both accuracy and MCC metrics. Among the traditional machine learning baselines, logistic regression and support vector machines performed comparatively well, yet neither matched the proposed framework's combination of high recall and low false classifications. Post-hoc tests confirmed that the proposed approach significantly outperformed all other models across accuracy, F1, balanced accuracy, and ROC scores, adjusting for multiple comparisons with Bonferroni correction (p < .05).

Taken together, these results demonstrate that the proposed framework outperforms both traditional supervised machine learning methods and direct zero-shot LLM classification for detecting both microaggressions (more subtle harmful content) and hate speech (more overt harmful content). The improvements in accuracy, F1 score, MCC, and ROC were especially pronounced, suggesting that pairwise textual comparisons, coupled with the Elo rating algorithm, can mitigate certain limitations imposed by content moderation while preserving robust classification performance.

framework consistently attained higher accuracy and recall than any other method, although average model performance in this task tended to be higher overall for all models. In a similar vein, when considering the more subtle microaggression detection problem, our approach performed better, showing its capacity to identify harmful communicative acts that have fewer explicit linguistic markers. These findings suggest that, when guided by the Elo rating algorithm, LLMs can score harmful communication with higher reliability and reduce some of the limitations of LLMs due to the existing content moderation systems (Cui et al., 2024; Devlin et al., 2019; Xie et al., 2024).

**Figure 3. Accuracy Comparisons for the Hate Speech Classification Task**

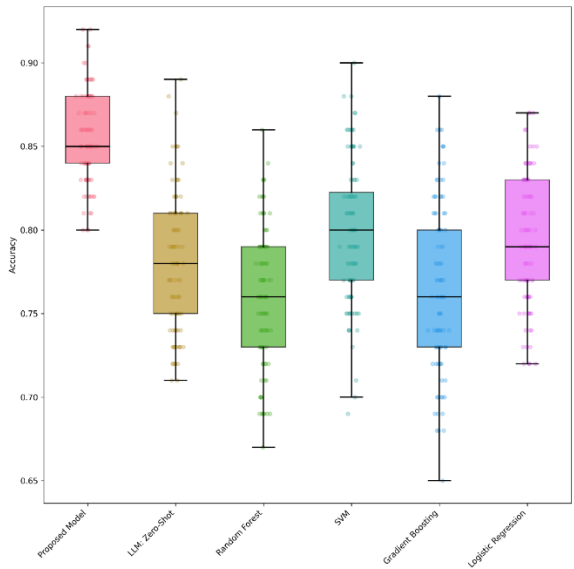

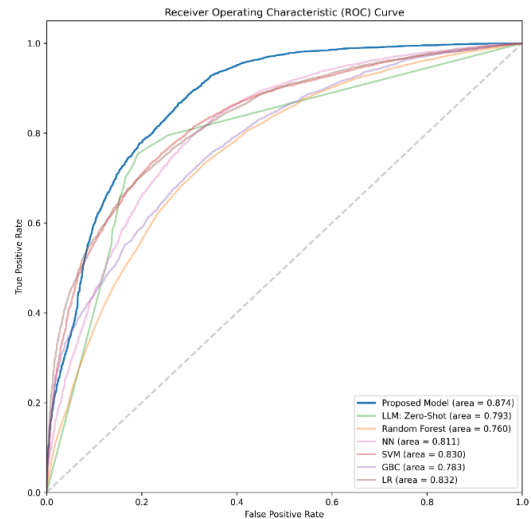

**Figure 4. ROC Curve for the Hate Speech Classification Task**

## 5. Discussion

In this study, we introduce a new method by combining LLMs with the Elo rating system to detect harmful content in both subtle (e.g., microaggressions) and overt (e.g., hate speech) forms. We find that our approach outperforms both traditional machine learning methods (e.g., random forests, support vector machines, gradient boosting machines, and logistic regression) and a zero-shot LLM prompting technique. In the hate speech detection task—which contained a more obvious linguistic signature for harmful content— the proposed

The implications of these findings are numerous. First, by using an adaptive, pairwise comparison framework, our model directly addresses one of the key challenges of harmful content detection, producing reliable analyses and adhering to content moderation systems. (Gillespie, 2018). Unlike existing strategies that use explicit classification prompts or that need large, labeled datasets (Brown et al., 2020), our approach merely asks the LLM for simple binary comparisons, which reduces the probability of triggering overly cautious filtering mechanisms (Cui et al., 2024). Second, this method eases some of the major limitations associated with manual coding and traditional machine-learning procedures, such as high labor costs and large data requirements (Krippendorff, 2012; Schmidt & Wiegand, 2017). Instead of collecting and labeling massive datasets, organizations can now incrementally incorporate new material into the Elo system, thereby making the detection process for emerging hateful lexicons or subtle forms of aggression more efficient (Weidinger et al., 2022). Third, from an organizational research point of view, the option to rank texts on a probabilistic scale instead of using rigid binary judgments allows for more fine-grained interventions and opportunities to study such phenomena, which in turn enables managers and researchers to address harmful communication in many nuanced ways.

## 6. Limitations and Future Directions

Despite the encouraging outcomes observed in this study, several limitations warrant discussion. First, although the use of our balanced incomplete block design allows for higher data coverage compared to a fully saturated design, using this design still amounts to a large number of pairwise comparisons per entry, which is resource-demanding (Montgomery, 2017). This can be more problematic with the larger data sets. Second, we provided our empirical demonstration by testing the performance of our framework with the GPT-4o-mini model (OpenAI, 2024); future work should examine the extent to which larger or other LLM architectures produce better results or produce different results entirely. Third, while our method delivers a data-efficient solution that is highly relevant for applications in social science contexts, where smaller and more specialized data sets are common, traditional machine learning algorithms may still be superior in large-scale settings where extensive computational resources and carefully curated training data with labels are available (Brown et al., 2020). Consequently, this is an open question regarding the generalizability of our findings to extremely large datasets, and more work is needed to determine the extent to which the Elo-based LLM classification approach can outperform other methods in high-data contexts.

Moving forward, exploring alternative or advanced versions of the Elo rating system—such as Bayesian Elo or adaptive Elo—holds promise for enhancing the flexibility and responsiveness of harmful content detection, particularly in rapidly evolving linguistic environments (Pelánek, 2016). Bayesian Elo can, for example, integrate priors over the estimated probabilities, whereas adaptive Elo variants can incrementally update ratings in near real-time and capture subtle changes in language usage while requiring fewer pairwise comparisons. Future research may also expand the reach of this pairwise comparison approach beyond harmful communication to other organizational classification tasks, including sentiment, deception, and conflict detection. Such expansions would shed light on the robustness and generalizability of the method while potentially reducing dependence on large, labeled datasets—thereby benefiting domains where data collection is both costly and ethically sensitive.

## 7. Conclusion

By integrating large language models with an Elo-based comparative framework, this study advances a more resource-efficient and adaptive method for detecting both subtle and overt harmful communication. Our results demonstrate the promise of using pairwise comparisons to overcome content moderation constraints without sacrificing classification accuracy. While offering direct benefits for research and practice in organizational research, we hope our approach can facilitate other applications across other social sciences that require nuanced textual analysis.